# A Hybrid Framework of Vision Transformer and Gated Recurrent Unit for Detection of Mosquito Diseases

Danial Sharifrazi
Institute for Intelligent Systems Research and Innovations (IISRI), Deakin University, Geelong, Australia
d.sharifraz@deakin.edu.au

Saadat Behzadi
Department of Electronic Engineering, University of Bologna, Bologna, Italy
saadat.behzadi@studio.unibo.it

Nouman Javed
Institute for Intelligent Systems Research and Innovations (IISRI), Deakin University, Geelong, Australia
n.javed@deakin.edu.au

Roohallah Alizadehsani
Institute for Intelligent Systems Research and Innovations (IISRI), Deakin University, Geelong, Australia
r.alizadehsani@deakin.edu.au

Prasad N. Paradkar
CSIRO Health and Biosecurity, Australian Animal Health Laboratory, Geelong, Australia
prasad.paradkar@csiro.au

Asim Bhatti
Institute for Intelligent Systems Research and Innovations (IISRI), Deakin University, Geelong, Australia
asim.bhatti@deakin.edu.au

***Abstract*— Identifying dengue virus-infected mosquitoes from control mosquitoes is a major challenge in analyzing mosquito locomotion behavior due to the small size and complexity of the video background. Conventional AI methods are often unable to extract accurate features from video frames and produce erroneous features. In this study, a three-step framework is introduced: first, mosquitoes are identified and the background is removed using the YOLO 11M model, then visual features are extracted using the Vision Transformer (ViT), and finally the videos are classified with a convolutional GRU (ConvGRU) classifier. A comparative analysis of different models, including Recurrent Neural Network (RNN), Long Short-Term Memory (LSTM), Gated Recurrent Unit (GRU), and their convolutional versions showed that the ConvGRU model achieved the best performance; it achieved 88.88% accuracy, 84.45% precision, 82.82% recall, and 82.81% F1 score. These results demonstrate that combining convolutional models with sequence-based networks, especially in the ConvGRU model, allows the simultaneous extraction of precise spatial features and long-term temporal dependencies from mosquito movements. Finally, the proposed framework provides a reliable solution for analyzing mosquito behavior in complex environments.**

***Keywords—Computer Vision, Vision Transformer, Video Classification, Object Detection***

## I. INTRODUCTION

In recent years, one of the major challenges of AI-based machine learning has been the complex behavioral analysis of small creatures in video, especially when the size of the objects is very small and their movements are irregular and fast. Identifying infected mosquitoes is a case in point; these mosquitoes play a key role in transmitting viruses such as dengue and Zika, infecting millions of people annually [1].

Traditional monitoring methods, including morphological measurements, are unable to detect behavioral changes caused by infection, as these changes usually occur at the level of subtle and long-term movements and are not detectable by conventional tools [2]. Biological studies indicate that viruses can modulate the neural activity and movement patterns of mosquitoes, leading to movements that are not detectable by direct observation or traditional methods [3].

Recent advances in machine vision and deep learning have shown that deep neural networks can be used to identify and track small objects in videos [4]. For example, methods such as Mask Region-Based Convolutional Neural Network (Mask R-CNN) and Convolutional Neural Network (CNN)-based systems have been able to perform very well in detecting eggs and tracking mosquito movements in complex environments [5, 6]. Also, the use of density map-based models and multiscale optical flow has made it possible to extract collective movement patterns and reduce background noise and disturbances [7]. These achievements show that integrating spatiotemporal movement information with deep learning networks can extract meaningful information from the collective behavior of mosquitoes. However, there are significant limitations: mosquitoes in the images occupy a small part of the frame and a large part of the pixels correspond to the fixed background such as the cage or ambient lighting. This condition causes CNN networks to focus incorrectly on background features and ignore real behavioral patterns [8]. Furthermore, detecting behavioral changes due to infection requires long-term motion analysis and spatiotemporal data aggregation, which many traditional CNN models are unable to extract [2].

Some attempts to overcome these limitations have included the use of a combination of object recognition methods and sequential models. For example, systems that have used motion tracking with CNN and optical flow have been able to capture behavioral changes associated with infection [6]. Also, density map-based models have enabled the compression of spatiotemporal information and the reduction of noise effects [9, 10]. Despite these advances, challenges remain in removing background noise and preserving long-term motion features [10].

To overcome the existing limitations, in this study, we propose a two-stage framework. In the first stage, using the YOLO-V11M network, the location of mosquitoes in each frame is identified and unnecessary parts of the background are removed; so that finally only the identified mosquitoes remain and the entire background is completely dark [11, 12]. As seen in Fig. 1, this process (from detection in a crowded environment to creating the final mask) ensures that the model focuses only on mosquitoes and minimizes interference from environmental noise. Then, the extracted data is fed to a Vision Transformer-based image feature extractor to extract meaningful image features from each frame. After that, these features are fed to a Convolutional GRU-based sequential classifier to model the complex and long-term patterns of mosquito movement.

In addition to preserving meaningful features and spatio-temporal movement patterns, this method also reduces the effect of noise and background. Also, the use of sequential networks makes the proposed framework have good generalization ability even with a limited number of labeled samples. The aim of this study is to present an effective and generalizable approach for analyzing videos containing small objects in biomedical fields, so that in addition to high accuracy, it also has biological interpretability and can be generalized to other similar problems in small object analysis. This method not only reduces the limitations of previous methods, but also provides a practical and usable framework for analyzing collective movements of small organisms in complex environments.

The rest of this paper is organized as follows: First, in the proposed methodology section, the details of the developed framework, including the target-centric preprocessing approach, background removal and insect localization, feature extraction using the ViT model, and finally temporal classification using the ConvGRU network will be fully explained. Next, the performance of the proposed model is compared with other methods and the impact of selecting different components (such as YOLO versions and feature extractors) is evaluated in the results section. Finally, our work is concluded in the conclusions section by reviewing the achievements and examining the significance of the research findings.

## II. DATASET

The data collection process in this study was carried out in a cubic cage with the presence of 15 mosquitoes, which were provided with a source of sugar water inside the cage to maintain their survival and nutrition. The behavioral images of the mosquitoes were recorded continuously by a camera placed in front of the cage over a period of 1 to 13 days to accurately record all their activities during both day and night. The statistical population of this study included non-transmitting mosquitoes as well as vectors of dengue and Zika viruses, which ultimately led to the formation of a data set in 3 separate classes; these classes included dengue-infected mosquitoes, Zika-infected mosquitoes, and the control group (uninfected).

According to the examples presented in Fig. 2, several environmental challenges were observed during the experiment that can make the feature extraction process difficult. These factors include light fluctuations between day and night, uneven distribution of light on the left and right sides of the cage, and variations in imaging angles. The presence of such interfering variables increases the risk of creating false patterns in the data, which will ultimately lead to a decrease in the accuracy of the model and a weakening of the validity of the classification results.

## III. PROPOSED METHODOLOGY

In this research, we aim to classify videos of mosquitoes flying. One of the main challenges in this case study is the small size of mosquitoes in the video frames. Therefore, mosquitoes occupy only such a limited part of the video pixels. In this case study, direct feature extraction often leads to background feature extraction. For this reason, we use a target-focused approach in the preprocessing section.

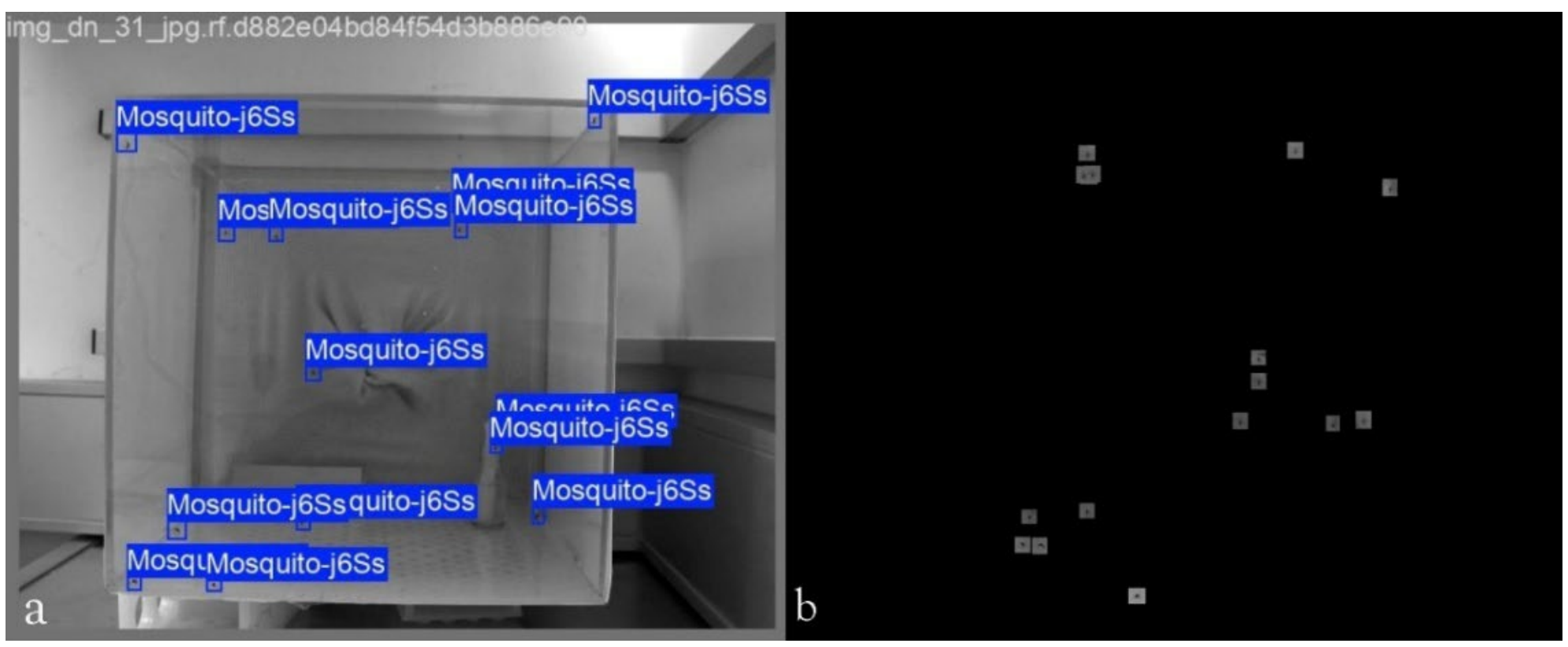

Fig. 1: Target localization and background removal process; (a) Mosquito detection in complex cage environment using YOLO-V11M model, (b) final masked frame where only target subjects are retained for feature extraction and the background is completely removed.

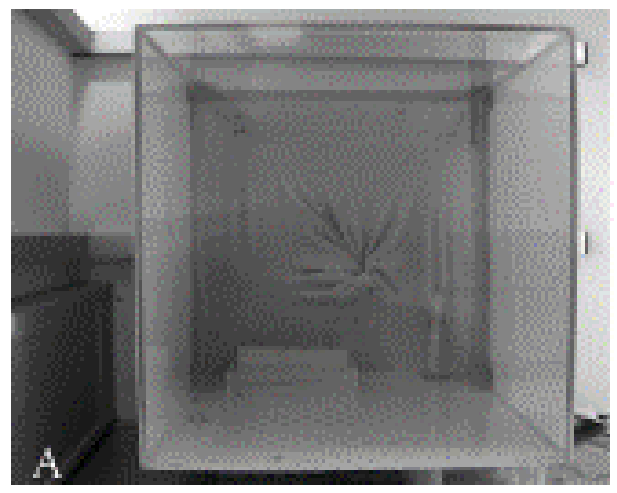
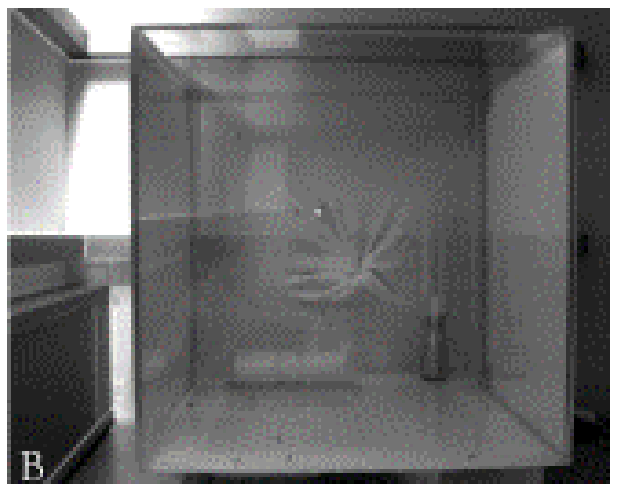

Fig. 2: Two sample frames of recorded videos regarding non-infected mosquitoes in A) daytime

### A. Background Elimination and Target Localization

The main goal in this section is to prevent feature extraction from the background. For this reason, we use a pre-trained YOLO model to localize insects in the video. Instead of training the object recognition model with all the video frames, we annotate only the initial few frames of the video and fine-tune the YOLO model with them. After detecting the location of insects in the frames, we set all image pixels except the detected boxes to zero. Thus, a mask is created that is able to filter insects from the frames and entirely remove the background.

### B. Frame-Level Feature Extraction

Following the creation of the background mask, all frames are resized to 224 × 224 and normalized between 0 and 1. If a video does not have enough frames, it is discarded from the process, as it cannot properly reflect the flight behavior of mosquitoes.

To have a robust representation of the frames, we use a ViT model with frozen layers, pre-trained with ImageNet weights, as a feature extractor. To avoid overfitting, the ViT model does not receive any fine-tuning. Each frame is independently entered into the ViT, and higher-level features are extracted from the frames.

Since videos may have different lengths and therefore different numbers of frames, a padding technique is applied. In this way, the length of all features obtained from the videos is the same.

### C. Temporal Classification

In order for the temporal features of the videos to be effectively learned, a classifier that is able to be trained using these temporal features is implemented. Specifically, a ConvGRU model is used. The proposed ConvGRU is integrated with a global average pooling layer and a dropout layer to reduce the computational load and prevent overfitting. Finally, a softmax layer classifies the high-level features produced by the ConvGRU into three classes of non-infected, dengue-infected, and Zika-infected videos.

To ensure the accuracy of the results obtained, the proposed model is evaluated through five-fold cross-validation. Furthermore, the outcomes of all folds are presented in the results section through the mean and standard deviation (std). The proposed method and similar models are evaluated using accuracy, precision, recall, and F1-score metrics to ensure a reliable classification. Table I describes the whole proposed framework briefly step by step.

TABLE I: PSEUDOCODE OF THE PROPOSED METHODOLOGY

**Algorithm:** Target-Focused Video Classification Framework

**Input:** Video dataset with three classes: Control, DENV2, ZIKV
**Output:** Classification performance metrics

```
1:  for each video in the dataset do:
2:      Extract video frames
3:      Localize mosquito regions using fine-tuned YOLO detector
4:      Eliminate background by zeroing non-detected pixels
5:      Resize frames to 224 × 224 and normalize to [0, 1]
6:      Discard videos with insufficient temporal length
7:  end for

8:  Extract frame-level features using frozen ImageNet-pretrained ViT
9:  Pad feature sequences to a uniform temporal length

10: for each fold in five-fold cross-validation do:
11:     Partition data into training, validation, and test sets
12:     Train temporal classifier (ConvGRU with pooling and dropout)
13:     Infer class labels for test samples
14:     Evaluate performance using accuracy, precision, recall, and F1-score
15: end for

16: Report mean and standard deviation of metrics across folds
```

## IV. RESULTS

### A. Comparative Analysis of Model Performance

The results of the comparative analysis of the performance of different models in detecting dengue virus-infected mosquitoes and control (uninfected) mosquitoes are presented in Fig. 3. This figure shows the performance of the proposed model compared to several baseline architectures including RNN, LSTM, GRU and their convolutional versions (ConvRNN, ConvLSTM and ConvGRU) based on the metrics of Accuracy, Precision, Recall and F1-score.

As can be seen in Fig. 3, sequential models did not perform well in solving this problem. More precisely, the RNN model achieved an accuracy of 75.55% and the LSTM model achieved an accuracy of approximately 70%, indicating that these two models are unable to extract complex and long-term movement patterns from mosquito videos. Although the GRU model outperformed the RNN and LSTM by recording an accuracy of 83.33%, it still lags behind the convolutional models.

By using convolutional versions of sequence-based models, significant improvements are observed in all evaluation criteria. By more effectively preserving and exploiting the spatial structure in the extracted feature maps while simultaneously modeling temporal dependencies, the ConvRNN and ConvLSTM models have been able to achieve higher accuracy and stability than their non-convolutional versions. These results indicate that in the analysis of videos containing small objects, preserving the spatial organization of features in the temporal modeling stage plays an important role in increasing the recognition quality.

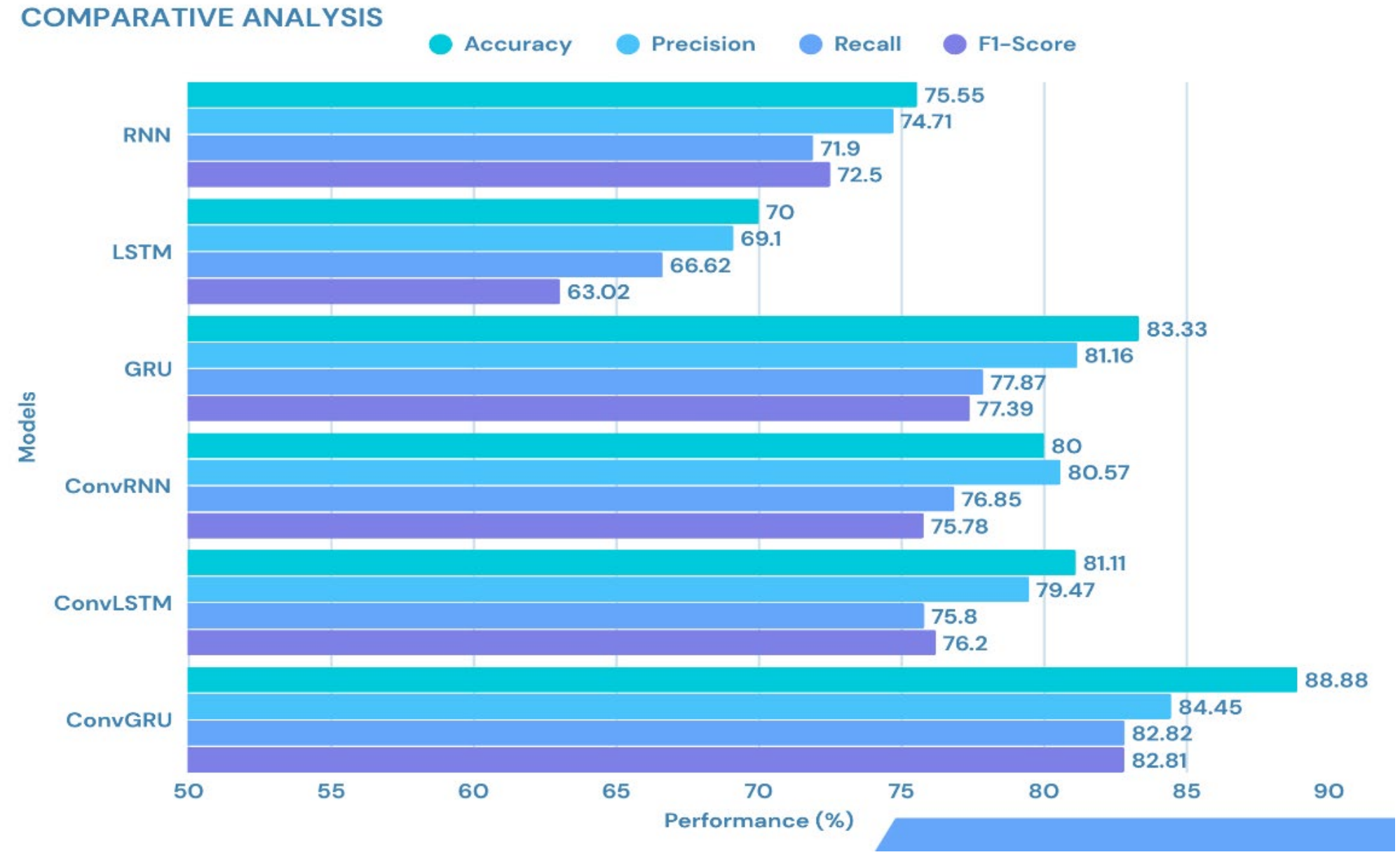

Fig. 3: Comparative analysis of the performance of different models in detecting infected and uninfected mosquitoes

Among all the models examined, the ConvGRU model showed the best performance. As can be seen in Fig 3, this model significantly outperforms other methods, achieving an accuracy of 88.88%, a precision of 84.45%, a recall of 82.82%, and an F1-score of 82.81%. This superiority reflects the high ability of ConvGRU to simultaneously learn fine spatial features and long-term temporal dependencies in mosquito movement patterns.

Overall, the results of this comparative analysis show that the use of convolutional-sequential architectures, especially ConvGRU, is much more effective than classical sequential models for detecting subtle behavioral differences in videos containing very small and moving objects in complex environments. These findings confirm the selection of the ConvGRU model as the core of the proposed framework for the mosquito infestation detection problem.

### B. *Ablation study*

In this section, we conducted various experiments to select the best model components to evaluate the impact of each on the final performance, which are explained in detail below.

#### *1) Preprocessing*

For the preprocessing stage, several YOLO models were used and after evaluating their performance, it is found that the YOLO 11M model performed best. The results of this experiment are given in Table II.

#### *2) Feature Extraction*

In this stage, several different feature extraction methods were also tested and concluded that ViT has the best performance in feature extraction. The results of these experiments are shown in Table III. To evaluate the results, four evaluation metrics were used: Accuracy, Precision, Recall, and F1-Score. Since ViT is based on transform-based models and has an attention-based feature, it was able to extract features from images well.

TABLE II: DIFFERENT YOLO VERSIONS PERFORMANCE BASED ON MAP50.

| Yolo Version | mAP50 (%) |
|---|---|
| YOLO 8N | 96.9 |
| YOLO 8M | 96.7 |
| YOLO 11N | 96.5 |
| **YOLO 11M** | **97.8** |

TABLE III: PERFORMANCE COMPARISON OF VARIOUS FEATURE EXTRACTION MODELS

| Feature Extractor | Accuracy | Precision | Recall | F1-Score |
|---|---|---|---|---|
| **ResNet** | 0.2111± 0.0889 | 0.0704 ± 0.0296 | 0.3333 ± 0.0000 | 0.1131 ± 0.0428 |
| **VGG19** | 0.2667 ± 0.1379 | 0.0889 ± 0.0460 | 0.3333 ± 0.0000 | 0.1339 ± 0.0599 |
| **EfficientNet** | 0.2111 ± 0.0889 | 0.0704 ± 0.0296 | 0.3333 ± 0.0000 | 0.1131 ± 0.0428 |
| **ViT** | 0.8889 ± 0.0351 | 0.8446 ± 0.1317 | 0.8283 ± 0.1067 | 0.8282 ± 0.1145 |

In contrast, CNN-based models that lacked attention-based features and also had completely blackened image backgrounds were unable to extract appropriate features

### 3) Hyperparameter Tuning

To tune the hyperparameters of the model, the Optuna platform was utilized [13]. This platform uses the Bayesian Optimization method to search the hyperparameter space. To do this, different ranges of hyperparameters were given to this platform and obtained the best hyperparameters. The parameters used for this task are: Number of Neurons, Dropout Rate, Learning Rate, Batch Size, and Optimizer. The range of parameters and their best tuning results are shown in Table IV.

### 4) Classifier Selection

To select the best classifier, various models, including RNN, GRU, LSTM, CNN-RNN, CNN-LSTM, and CNN-GRU models, were tested. After reviewing the results, it was found that the CNN-GRU model performed the best.

Finally, through these experiments and ablation study, we concluded that the combination of YOLO 11M for mosquito detection, ViT for feature extraction, and CNN-GRU as a classifier is the best combination for analyzing mosquito behavior.

## V. DISCUSSION

In this study, a three-step framework for analyzing mosquito behavior in videos containing very small objects was presented. The results show that focusing directly on the target area and effectively removing the background plays an important role in improving the classification accuracy and reducing the biases caused by light and environmental conditions. This feature allows the model to focus on the real behavior of mosquitoes instead of learning background-dependent patterns.

Another advantage of the proposed method is its ability to model long-term movement patterns. The use of sequential structures allows for the extraction of gradual and cumulative changes in movement paths, which is also biologically important for detecting the effect of viral infections. Also, the stable performance of the model in low-data conditions indicates that the proposed framework is suitable for data-limited biomedical applications.

TABLE IV: HYPERPARAMETER SEARCH SPACE AND OPTIMAL CONFIGURATION

| *Name of Hyperparameters* | *Range of Hyperparameters* | *Best Outcome* |
|---|---|---|
| **Number of Neurons** | [32, 64, 128, 256] | 128 |
| **Dropout Rate** | [0.1 – 0.5] | 0.25 |
| **Learning Rate** | [1e-5 – 1e-3] | 1e-5 |
| **Batch Size** | [16, 32, 64, 128] | 16 |
| **Optimizer** | [Adam, SGD, RMSprop] | Adam |

Despite these advantages, this study also has some limitations. The evaluation of the method was conducted only on a specific dataset of mosquitoes, and data from other insect species or different environmental conditions were not available. Consequently, a more comprehensive examination of the generalizability of the model to more diverse scenarios requires the collection and evaluation on larger datasets in the future.

Overall, the findings of this study show that intelligent background removal and focusing on movement patterns is an efficient solution for video analysis of very small subjects; an approach that has great potential for generalization to other similar challenges in the field of biological behavior analysis.

## VI. CONCLUSION

In this study, we focused on solving a straightforward but often overlooked challenge: analyzing the movements of very small mosquitoes in videos with complex backgrounds. Our results showed that conventional AI methods are unable to provide optimal performance in this area due to the neglect of spatial features and long-term temporal dependencies. Our proposed framework, by integrating the YOLO 11M, ViT, and ConvGRU models, showed that it is quite possible to extract accurate features that are aligned with the biological structure of mosquito movements. Applying the ConvGRU model in this framework not only led to a significant improvement in recognition accuracy, but also provided much more stable results compared to standard sequence-based approaches. Overall, the achievement of this research is significant both from the perspective of methodological innovation and from the perspective of application potential, and shows that combining advanced machine learning techniques with biological video analysis can be a reliable path to a deeper and more precise understanding of subtle behaviors in small organisms.

## APPENDIX

Animal ethics and consent to participate declarations are not applicable for this research as it did not involve humans or animals. The source code of the research is available via the link below:

https://github.com/danialsharifrazi/Video-Classification-VisionTransformer-ConvGRU